\documentclass{article}
\usepackage{spconf,amsmath,graphicx}
\usepackage{amssymb,cite,multirow}

\usepackage{subcaption}

\usepackage{standalone}

\usepackage{tabularx}
\newcolumntype{L}[1]{>{\raggedright\let\newline\\\arraybackslash\hspace{0pt}}m{#1}}
\newcolumntype{C}[1]{>{\centering\let\newline\\\arraybackslash\hspace{0pt}}m{#1}}
\newcolumntype{R}[1]{>{\raggedleft\let\newline\\\arraybackslash\hspace{0pt}}m{#1}}

\usepackage{siunitx}
\usepackage[acronym]{glossaries}
\usepackage{balance}
\usepackage{hyperref}

\usepackage[english,capitalize]{cleveref}
\usepackage{pgfplots}
\usepackage{tikz}
\usetikzlibrary{positioning,chains,calc,shapes.geometric, shadows, shapes.misc}
\usetikzlibrary{fit}
\usetikzlibrary{decorations.pathreplacing,angles,quotes}

\newacronym{AM}{AM}{acoustic model}
\newacronym{ASR}{ASR}{Automatic Speech Recognition}
\newacronym{BF}{BF}{beamforming}
\newacronym{BAN}{BAN}{Blind Analytical Normalization}
\newacronym{CNN}{CNN}{Convolutional Neural Network}
\newacronym{EM}{EM}{Expectation Maximization}
\newacronym{TDNN-F}{TDNN-F}{factorized TDNN}
\newacronym{GMM}{GMM}{Gaussian Mixture Model}
\newacronym{GSS}{GSS}{Guided Source Separation}
\newacronym{LF-MMI}{LF-MMI}{Lattice-Free Maximum Mutual Information}
\newacronym{MM}{MM}{Mixture Model}
\newacronym{MVDR}{MVDR}{Minimum Variance Distortionless Response}
\newacronym{NN}{NN}{neural network}
\newacronym{RIR}{RIR}{room impulse response}
\newacronym{TDNN}{TDNN}{Time Delay Neural Network}
\newacronym{STFT}{STFT}{Short Time Fourier Transform}
\newacronym{WER}{WER}{word error rate}
\newacronym{WPE}{WPE}{Weighted Prediction Error}

\title{An Investigation into the Effectiveness of Enhancement \\in ASR Training and Test for CHiME-5 Dinner Party Transcription}

\name{Catalin Zorila$^{1}$, Christoph Boeddeker$^{2}$, Rama Doddipatla$^{1}$ and Reinhold Haeb-Umbach$^{2}$
\thanks{\copyright 2019 IEEE. Accepted for ASRU 2019.}}
\address{
$^{1}$ Toshiba Cambridge Research Laboratory, Cambridge, United Kingdom\\
$^{2}$ Paderborn University, Department of Communications Engineering, Paderborn, Germany}
\begin{document}
\ninept
\maketitle
\begin{abstract}
Despite the strong modeling power of neural network acoustic models, speech enhancement has been shown to deliver additional word error rate improvements if multi-channel data is available.
However, there has been a longstanding debate whether enhancement should also be carried out on the ASR training data.
In an extensive experimental evaluation on the acoustically very challenging CHiME-5 dinner party data we show that: (i) cleaning up the training data can lead to substantial error rate reductions, and (ii) enhancement in training is advisable as long as enhancement in test is at least as strong as in training.
This approach stands in contrast and delivers larger gains than the common strategy reported in the literature to augment the training database with additional artificially degraded speech.
Together with an acoustic model topology consisting of initial CNN layers followed by factorized TDNN layers we achieve with \SI{41.6}{\%}  and \SI{43.2}{\%} WER on the DEV and EVAL test sets, respectively,  a new single-system state-of-the-art result on the CHiME-5 data. This  
is a \SI{8}{\%} relative improvement compared to the best word error rate published so far for a speech recognizer without system combination.
\end{abstract}
\section{Introduction}
\label{sec:intro}

Neural networks have outperformed earlier \gls{GMM} based acoustic models in terms of modeling power and increased robustness to acoustic distortions.
Despite that, speech enhancement  has been shown to deliver additional \gls{WER} improvements, if multi-channel data is available.
This is due to their ability to exploit spatial information, which is reflected by phase differences of microphone channels in the \gls{STFT} domain. This information is not accessible by the \gls{ASR} system, at least not if it operates on the common log mel spectral or cepstral feature sets.
Also, dereverberation algorithms have been shown to consistently improve ASR results, since the temporal dispersion of the signal caused by reverberation is difficult to capture by an ASR acoustic model~\cite{Delcroix2014LINEARPD}.

However, there has been a long debate whether it is advisable to apply speech enhancement on data used for ASR training, because it is generally agreed upon that the recognizer should be exposed to as much acoustic variability as possible during training, as long as this variability matches the test scenario~\cite{Bippus_is99, Baker_asru2015, Vincent_csl2016}.
Multi-channel speech enhancement, such as acoustic \gls{BF} or source separation, would not only reduce the acoustic variability, it would also result in a reduction of the amount of training data by a factor of $M$, where $M$ is the number of microphones~\cite{Menne_chime4_2016}. Previous studies have shown the benefit of training an ASR on matching enhanced speech~\cite{Deng_isclp2000, Delcroix_is2013} or on jointly training the enhancement and the acoustic model~\cite{Li_is2016}.
Alternatively, the training data is often artificially increased by adding even more degraded speech to it.
For instance, Ko et al.~\cite{Ko_icassp17} found that adding simulated reverberated speech improves accuracy significantly on several large vocabulary tasks.
Similarly, Manohar et al.~\cite{Manohar_icassp19} 
improved the \gls{WER} of the baseline CHiME-5 system by relative \SI{5.5}{\%} by augmenting the training data with approx. \SI{160}{hrs} of simulated reverberated speech. 
However, not only can the generation of new training data be costly and time consuming, the training process itself is also prolonged if the amount of data is increased.

In this contribution we advocate for the opposite approach. Although we still believe in the argument that ASR training should see sufficient variability, instead of adding degraded speech to the training data, we clean up the training data. We make, however, sure that the remaining acoustic variability is at least as large as on the test data. By applying a beamformer to the multi-channel input, we even reduce the amount of training data significantly. Consequently, this leads to cheaper and faster acoustic model training.

We perform experiments using data from the CHiME-5 challenge which focuses on distant multi-microphone conversational ASR in real home environments~\cite{Barker2018CHiME5}.
The CHiME-5 data is heavily degraded by reverberation and overlapped speech. As much as \SI{23}{\%} of the time more than one speaker is active at the same time~\cite{Zorila_icassp19}.
The challenge's baseline system poor performance (about~\SI{80}{\%}~\gls{WER}) is an indication that ASR training did not work well.
Recently, \Gls{GSS} enhancement on the test data was shown to significantly improve the performance of an acoustic model, which had been trained with a large amount of unprocessed and simulated noisy data~\cite{Kanda2019}.
\Gls{GSS} is a spatial mixture model based blind source separation approach which exploits the annotation given in the CHiME-5 database for initialization and, in this way, avoids the frequency permutation problem~\cite{Boeddecker2018chime5}.

We conjectured that cleaning up the training data would enable a more effective acoustic model training for the CHiME-5 scenario. 
We have therefore experimented with enhancement algorithms of various strengths, from relatively simple beamforming over single-array \gls{GSS} to a quite sophisticated multi-array \gls{GSS} approach,  and tested all combinations of training and test data enhancement methods.
Furthermore, compared to the initial \Gls{GSS} approach in~\cite{Boeddecker2018chime5}, we describe here some modifications, which led to improved performance.
We also propose an improved neural acoustic modeling structure compared to the CHiME-5 baseline system described in \cite{Manohar_icassp19}. It consists of initial \gls{CNN} layers followed by \gls{TDNN-F} layers, instead of a homogeneous \gls{TDNN-F} architecture.

Using a single acoustic model trained with \SI{308}{hrs} of training data, which resulted after applying multi-array GSS data cleaning and a three-fold speed perturbation, we achieved a \gls{WER} of \SI{41.6}{\%} on the development (DEV) and \SI{43.2}{\%} on the evaluation (EVAL) test set of CHiME-5, if the test data is also enhanced with multi-array GSS. 
This compares very favorably with the recently published top-line in \cite{Kanda2019}, where the single-system best result, i.e., the WER without system combination, was \SI{45.1}{\%} and \SI{47.3}{\%} on DEV and EVAL, respectively, using an augmented training data set of \SI{4500}{hrs} total.

The rest of this paper is structured as follows. Section~\ref{sec:c5} describes the CHiME-5 corpus, Section~\ref{sec:gss} briefly presents the guided source separation enhancement method, Section~\ref{sec:exp} shows the ASR experiments and the results, followed by a discussion in Section~\ref{sec:disc}. Finally, the paper is concluded in Section~\ref{sec:concl}.

\section{CHiME-5 corpus description}
\label{sec:c5}

The CHiME-5 corpus comprises twenty dinner party recordings (sessions) lasting for approximately \SI{2}{hrs} each. A session contains the conversation among the four dinner party participants.  
Recordings were made in kitchen, dining and living room areas with each phase lasting for a minimum of \SI{30}{mins}.
16 dinner parties were used for training, 2 were used for development, and 2 were used for evaluation. 

There were two types of recording devices collecting CHiME-5 data: distant 4-channels (linear)  Microsoft Kinect arrays (referred to as units or `U') and in-ear Soundman OKM II Classic Studio binaural microphones (referred to as worn microphones or `W'). Six Kinect arrays were used in total and they were placed such that at least two units were able to capture the acoustic environment in each recording area. Each dinner party participant wore in-ear microphones which were subsequently used to facilitate human audio transcription of the data.
The devices were not time synchronized during recording. Therefore, the W and the U signals had to be aligned afterwards using a correlation based approach provided by the organizers.
Depending on how many arrays were available during test time, the challenge had a single (reference) array and a multiple array track. For more details about the corpus, the reader is referred to~\cite{Barker2018CHiME5}.

\section{Guided source separation}
\label{sec:gss}

\gls{GSS} enhancement is a blind source separation technique originally proposed in \cite{Boeddecker2018chime5}\footnote{\url{https://github.com/fgnt/pb_chime5}} to alleviate the speaker overlap problem in CHiME-5.
Given a mixture of reverberated overlapped speech, \gls{GSS} aims to separate the sources using a pure signal processing approach. An \gls{EM} algorithm estimates the parameters of a spatial mixture model and the posterior probabilities of each speaker being active are used for mask based beamforming.

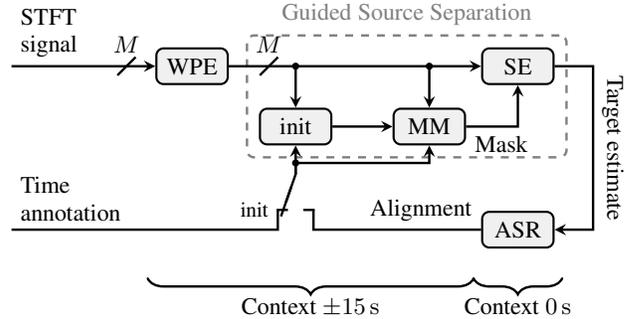
\begin{figure}[t]
	\centering
	\tikzset{%
  block/.style    = {draw, thick, rectangle, minimum height = 1.5em, minimum width = 3em, rounded corners=0.3em, fill=black!6},
  sum/.style      = {draw, circle, node distance = 2cm}, 
  cross/.style={path picture={\draw[black](path picture bounding box.south east) -- (path picture bounding box.north west)
		 (path picture bounding box.south west) -- (path picture bounding box.north east);}}
               }
\tikzstyle{branch}=[{circle,inner sep=0pt,minimum size=0.3em,fill=black}]

\begin{tikzpicture}[auto, line width=0.1em, node distance = 1cm]

\tikzset{pics/.cd,
	pic switch/.style args={#1 times #2}{code={
			\tikzset{x=#1/2,y=#2/2}
			\coordinate (-north west) at (-1,1);
			\coordinate (-north east) at (1,1);
			\coordinate (-south west) at (-1,-1);
			\coordinate (-south east) at (1,-1);
			\coordinate (-north) at (0,1);
			\coordinate (-east) at (1,0);
			\coordinate (-south) at (0,-1);
			\coordinate (-west) at (-1,0);
			
			\draw [line cap=rect] (-1,1) -- (-1,0.5);
			\draw [line cap=rect] (-1,-1) -- (-1,-0.5);
			\draw [line cap=round] (1, 0) -- ($(1, 0)!2/3!(-1.3,0.8)$);
			\draw [line cap=rect] ($(1, 0)!1/3!(-1.3,0.8)$) -- (-1.3,0.8);
	}}
}

\tikzset{>=stealth}
\tikzstyle{arrow}=[{}-{>}]

\node[block, align=center, at={(2.4,0)}](WPE){WPE};
\node[block, align=center, anchor=north west](BSSinit) at ($(WPE.south east) + (0.4, -0.3)$) {init};

\node[block, align=center, anchor=west](BSS) at ($(BSSinit.east) + (0.8, 0)$) {MM};

\node[block, align=center, anchor=west](BF) at ($(BSS.south east |- WPE.east) + (0.2, 0)$) {SE};
\node[block, align=center] at ($(BSS -| BF) - (BF) + (0, -1.8em) + (BSS -| BF)$) (ASR){ASR};

\pic [rotate=90] (switch) at ($(ASR.west -| BSSinit)!0.!(ASR.west -| BSS) + (0, 0.5)$) {pic switch={1.5em times 1.5em}};

\node[fit=(BSSinit)(BSS)(BF), inner sep=0.5em, draw, dashed, gray, rounded corners=0.3em, shift={(0,0em)}] (gssBox) {};

\node[text=gray, above, anchor=south] at (gssBox.north) {Guided Source Separation};

\coordinate(Y) at (0, 0);
\coordinate(a) at ($(ASR -| Y)$);
\draw[arrow] (Y) node[above right, align=left]{STFT\\signal} -- node[strike out,draw, anchor=center,pos=0.8,-]{} node[above=0.06,pos=0.8]{\small{$M$}}(WPE);
\draw[arrow] (WPE) -| node[strike out,draw, pos=0.3,-, anchor=center]{} node[above=0.06, pos=0.3]{\small{$M$}} (WPE -| BSSinit) node[branch]{} -- (BSSinit);
\draw[arrow] (WPE -| BSSinit) -| (WPE -| BSS) node[branch]{} -- (BSS);
\draw[arrow] (WPE -| BSS) -- (BF);

\draw[arrow] (BSSinit) -- (BSS);
\draw[arrow] (BSS.east) node[below right]{Mask} -| (BF);

\draw[] (a) node[above right,align=left]{Time\\annotation} -| (switch-north west);
\draw[arrow] (BF) -- +(1,0) -- node[rotate=270, anchor=south]{Target estimate} ($(ASR) + (1,0)$) -- (ASR);
\draw[] (ASR.west) node[above left]{Alignment} -++ (-0.5, 0) -| (switch-south west);
\draw[arrow] (switch-east) -++ (0, 0.15) node[branch]{} -| (BSSinit);
\draw[arrow] (switch-east) -++ (0, 0.15) -| (BSS);
\node[left] at (switch-north west){\footnotesize{init}};

%

\coordinate(contextMid) at ($(ASR.south)!1/2!(BSS|-ASR.north) + (0,-1.5em)$);
\draw[decoration={brace,amplitude=6pt,mirror,raise=5pt},line cap=round,decorate]
($(WPE.west |- contextMid) + (-0.1, 0)$) -- node[below=9pt] {Context $\pm \SI{15}{\second}$} (contextMid);
\draw[decoration={brace,amplitude=6pt,mirror,raise=5pt},line cap=round,decorate]
(contextMid) -- node[below=9pt] {Context $\SI{0}{\second}$} ($(ASR.east |- contextMid) + (0.1, 0)$);

\end{tikzpicture}
	\caption{Overview of speech enhancement system with \acrfull{WPE} dereverberation, \acrfull{MM} estimation, Source Extractor (SE) and \acrfull{ASR}.}
	\label{fig:enhancement_block}
\end{figure}

An overview block diagram of this enhancement by source separation is depicted in \cref{fig:enhancement_block}. It follows the approach presented in~\cite{Kanda2019}, which was shown to outperform the baseline version.
The system operates in the \gls{STFT} domain and consists of two stages:  (1) a dereverberation stage, and (2) a guided source separation stage. For the sake of simplicity, the overall system is referred to as GSS for the rest of the paper.
Regarding the first stage, the multiple input multiple output version of the \gls{WPE} method was used for dereverberation ($M$ inputs and $M$ outputs)~\cite{Yoshioka2012GWPE,Drude2018naraWPE}\footnote{\url{https://github.com/fgnt/nara_wpe}} and, regarding the second stage, it consists of a spatial \gls{MM} \cite{Ito2016cACGMM} and a source extraction (SE) component.
The model has five mixture components, one representing each speaker, and an additional component representing the noise class. 

\begin{figure}[t]
	\centering
	\begin{tikzpicture}[x=1em, y=2ex, scale=1]

\tikzset{>=stealth}

\tikzset{every picture/.style={line width=0.1em}}

\tikzstyle{block}=[draw,text centered, text width=3em, minimum height=4ex] 
\tikzstyle{branch}=[{circle,inner sep=0pt,minimum size=0.3em,fill=black}]
\tikzstyle{box}=[rectangle, rounded corners, draw=black, line width=1pt, text width=2cm]

\tikzstyle{arrow}=[{}-{>}]

\tikzstyle{block}=[draw, text centered, minimum height=2ex, inner sep=0, outer sep=0, fill=black!30] 

\node (p01-1) [block, text width=3em, fill=blue!50] at (0, 0) {\footnotesize P01};
\node (p01-2) [block, text width=1.7em, anchor=west] at ($(p01-1.east) + (2, 0)$) {};

\node (p02-1) [block, text width=2.5em] at (7, -1) {\footnotesize P02};

\node (p03-1) [block, text width=3.1em] at (-0.2, -2) {\footnotesize P03};
\node (p03-2) [block, text width=1em, anchor=west] at ($(p03-1.east) + (3.5, 0)$) {};

\node (p04-1) [block, text width=3em] at (-4, -3) {\footnotesize P04};
\node (p04-2) [block, text width=0.6em, anchor=west] at ($(p04-1.east) + (1.5, 0)$) {};
\node (p04-3) [block, text width=2em, anchor=west] at ($(p04-2.east) + (2, 0)$) {};

\node (Noise) [block, text width=10em] at (0, -4) {\footnotesize Noise};



\node[anchor=east, inner sep=0.2em] (source-1) at (-6, 0) {\footnotesize class $k=1$};
\node[anchor=east, inner sep=0.2em] (source-2) at ($(source-1.east) + (0, -1)$) {\footnotesize class $k=2$};
\node[anchor=east, inner sep=0.2em] (source-3) at ($(source-2.east) + (0, -1)$) {\footnotesize class $k=3$};
\node[anchor=east, inner sep=0.2em] (source-4) at ($(source-3.east) + (0, -1)$) {\footnotesize class $k=4$};
\node[anchor=east, inner sep=0.2em] (source-5) at ($(source-4.east) + (0, -1)$) {\footnotesize class $k=0$};

\coordinate (time-left) at ($(source-1.west) + (0, -6.5)$);
\coordinate (time-right) at (p02-1.east |- time-left);

\draw [arrow, black] (time-left) -- (time-right) node[above left] {time};

\draw [purple] ($(time-left -| Noise.west) + (0, -0.2)$) -- ($(p01-1.north -| Noise.west) + (0, 0.2)$);
\draw [purple] ($(time-left -| Noise.east) + (0, -0.2)$) -- ($(p01-1.north -| Noise.east) + (0, 0.2)$);

\node () [purple, below, inner sep=0, outer sep=0.2em] at (time-left -| Noise) {Extended utterance};

\draw [blue] ($(time-left -| p01-1.west) + (0, -0.2)$) -- ($(p01-1.north -| p01-1.west) + (0, 0.2)$);
\draw [blue] ($(time-left -| p01-1.east) + (0, -0.2)$) -- ($(p01-1.north -| p01-1.east) + (0, 0.2)$);

\node () [blue, above, fill=white, inner sep=0, outer sep=0.2em] at (time-left -| Noise) {Original utterance};

\draw [{<}-{>}, black!60!green] ($(p01-1.north west)$) -- node[above] {Context} ($(p01-1.north -| Noise.west)$);
\draw [{<}-{>}, black!60!green] ($(p01-1.north east)$) -- node[above] {Context} ($(p01-1.north -| Noise.east)$);

\draw [white, ,line cap=rect, line width=0.2em] (p02-1.north east) -- (p02-1.south east);
\draw [white, ,line cap=rect, line width=0.2em] (p04-1.north west) -- (p04-1.south west);


%
%

\end{tikzpicture}
	\caption{
		Visualization of time annotations on a fragment of the CHiME-5 data.
		The grey bars indicate source activity,
		the inner vertical blue lines denote the utterance boundaries of a segment of speaker P01,
		and the outer vertical red lines the boundaries of the extended utterance, consisting of the segment and the ``context'',
		on which the mixture model estimation algorithm operates.
	}
	\label{fig:activity}
\end{figure}
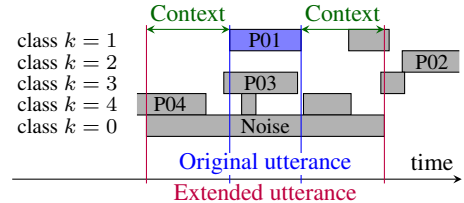

The role of the \gls{MM} is to support the source extraction component for estimating the target speech.
The class affiliations computed in the E-step of the EM algorithm are employed to estimate spatial covariance matrices of target signals and interferences, from which  the coefficients of an \gls{MVDR} beamformer are computed~\cite{Souden2010MVDR}.
The reference channel for the beamformer is estimated based on an SNR criterion\cite{Erdogan2016MVDR}. 
The beamformer is followed by a postfilter to reduce the remaining speech distortions ~\cite{Warsitz2007GEVBAN}, which in turn is followed by an additional (optional) masking stage to improve crosstalk suppression.
Those masks are also given by the mentioned class affiliations.
For the single array (CHiME-5) track, simulations have shown that multiplying the beamformer output with the target speaker mask improves the performance on the U data, but the same approach degrades the performance in the multiple array track~\cite{Boeddecker2018chime5}.
This is because the spatial selectivity of a single array is very limited in CHiME-5: the speakers' signals arrive at the array, which is mounted on the wall at some distance, at very similar impinging angles, rendering single array beamforming rather ineffective.
Consequently,  additional masking has the potential to improve the beamformer performance.
Conversely, the \gls{MM} estimates are more accurate in the multiple array case since they benefit from a more diverse spatial arrangement of the microphones, and the signal distortions introduced by the  additional masking rather degrade the performance.
Consequently, for our experiments we have used the masking approach for the single array track, but not for the multiple array one.

GSS exploits the baseline CHiME-5 speaker diarization information available from the transcripts (annotations) to determine when multiple speakers talk simultaneously (see~\cref{fig:activity}). This crosstalk information is then used to guide the parameter estimation of the \gls{MM} both during EM initialization (posterior masks set to one divided by the number of active speakers for active speakers' frames, and zero for the non-active speakers) and after each E-step (posterior masks are clamped to zero for non-active speakers).

The initialization of the \gls{EM} for each mixture component is very important for the correct convergence of the algorithm.
If the \gls{EM} initialization is close enough to the final solution, then it is expected that the algorithm will correctly separate the sources and source indices are not permuted across frequency bins. This has a major practical application, since frequency permutation solvers like \cite{Sawada2004Perm} become obsolete.

Temporal context also plays an important role in the EM initialization. Simulations have shown that a large context of 15 seconds left and right of the considered segment improves the mixture model estimation performance significantly for CHiME-5~\cite{Boeddecker2018chime5}. However, having such a large temporal context may become problematic when the speakers are moving, because the estimated spatial covariance matrix can become outdated due to the movement~\cite{Kanda2019}. Alternatively, one can run the EM first with a larger temporal context until convergence, then drop the context and re-run it for some more iterations. 
As shown later in the paper, this approach did not improve ASR performance. Therefore, the temporal context was only used for dereverberation and the mixture model parameter estimation, while  for the estimation of covariance matrices for beamforming the context was dropped and only the original segment length was considered~\cite{Kanda2019}.

Another avenue we have explored for further source separation improvement was to refine the baseline CHiME-5 annotations using ASR output (see \cref{fig:enhancement_block}).
A first-pass decoding using an \gls{ASR} system is used to predict silence intervals. Then this information is used to adjust the time annotations, which are used in the EM algorithm as described above.  When the ASR decoder  indicates silence  for a speaker, the corresponding class posterior in the \gls{MM} is forced to zero.

Depending on the number of available arrays for CHiME-5, two flavours of GSS enhancement were used in this work.
In the single array track, all 4 channels of the array are used as input ($M = 4$), and the system is referred to as GSS1.
In the multi array track, all six arrays are stacked to form a 24 channels super-array ($M = 24$), and this system is denoted as GSS6.
The baseline time synchronization provided by the challenge organizers was sufficient to align the data for GSS6.

\section{Experiments}
\label{sec:exp}

\subsection{General configuration}\label{ssec:setup}

Experiments were performed using the CHiME-5 data. Distant microphone recordings (U data) during training and/or testing were processed using the speech enhancement methods depicted in Table~\ref{tab:enh_meths}.
Speech was either left unprocessed, enhanced using a weighted delay-and-sum beamformer (BFIt)~\cite{Anguera_ieeetaslp2007} with or without dereverberation (WPE), or processed using the guided source separation (GSS) approach described in Section~\ref{sec:gss}. In Table~\ref{tab:enh_meths}, the strength of the enhancement  increases from top to bottom, i.e., GSS6 signals are much cleaner than the unprocessed ones.

The standard CHiME-5 recipes were used to: 
(i) train GMM-HMM alignment models, 
(ii) clean up the training data, and 
(iii) augment the training data using three-fold speed perturbation.
The acoustic feature vector consisted of 40-dimensional MFCCs appended with 100-dimensional i-vectors.
By default, the acoustic models were trained using the \gls{LF-MMI} criterion and a 3-gram language model was used for decoding~\cite{Barker2018CHiME5}.
Discriminative training (DT)~\cite{Ghoshal_is2013} and an additional RNN-based language model (RNN-LM)~\cite{Mikolov_is2010} were applied to improve recognition accuracy for the best performing systems.

\begin{table}[t]
\footnotesize
\caption{Naming of the speech enhancement methods.}
\begin{center}
\begin{tabular}{lcl}
	\hline
	Enhancement & Array & Label \\
	\hline
	 Unprocessed 															& Single/Multi  & None \\
	 BeamformIt~\cite{Anguera_ieeetaslp2007} 	& Single				& BFIt \\
	 WPE + BeamformIt~\cite{Manohar_icassp19} & Single 				& WPE+BFIt \\
	 WPE + GSS1 + BF w/o Context~\cite{Boeddecker2018chime5}	& Single & GSS1 \\
	 WPE + GSS6 + BF w/o Context~\cite{Boeddecker2018chime5} 	& Multi & GSS6 \\
	\hline
\end{tabular}
\end{center}
\label{tab:enh_meths}
\end{table}

\subsection{Acoustic model}

The initial baseline system~\cite{Barker2018CHiME5} of the CHiME-5 challenge uses  a \gls{TDNN} \gls{AM}.
However, recently it has been shown that introducing factorized layers into the TDNN architecture facilitates training deeper networks and also improves the ASR performance~\cite{Povey_is2018}.
This architecture has been employed in the new baseline system for the challenge~\cite{Manohar_icassp19}.
The \gls{TDNN-F} has 15 layers with a hidden dimension of 1536 and a bottleneck dimension of 160; each layer also has a resnet-style bypass-connection from the output of the previous layer, and a ``continuous dropout'' schedule~\cite{Manohar_icassp19}.
In addition to the \gls{TDNN-F}, the newly released baseline\footnote{\tiny\url{https://github.com/kaldi-asr/kaldi/tree/master/egs/chime5/s5b}} also uses simulated reverberated speech from worn microphone recordings for augmenting the training set, it employes front-end speech dereverberation and beamforming (WPE+BFIt), as well as robust i-vector extraction using 2-stage decoding.

\Glspl{CNN} have been previously shown to improve ASR robustness~\cite{AbdelHamid_ieeetaslp2014}. Therefore, combining \gls{CNN} and \gls{TDNN-F} layers is a promising approach to improve the baseline system of~\cite{Manohar_icassp19}.
To test this hypothesis, a CNN-TDNNF \gls{AM} architecture\footnote{\tiny\url{https://github.com/kaldi-asr/kaldi/tree/master/egs/swbd/s5c}}
consisting of 6 CNN layers followed by 9 TDNN-F layers was compared against an AM having 15 TDNN-F layers.
All \gls{TDNN-F} layers have the topology described above.

\begin{table}[t]
\footnotesize
\caption{Comparison of baseline \gls{TDNN-F}~\cite{Manohar_icassp19} and proposed CNN-TDNNF~\gls{AM}s in terms of \gls{WER} for the DEV (EVAL) set.}
\begin{center}
	\setlength{\tabcolsep}{5pt}  
\begin{tabular}{llccc}
	\hline
	 AM 	& 	Enh. in trng / \si{hrs} 	& 	Enh. in test  	& 	WER (\si{\%}) \\
	\hline
	TDNNF~\cite{Manohar_icassp19} & None / $1416$ &  WPE+BFIt & $69.6$ ($61.7$) \\
	CNN-TDNNF 										& None / $1416$ 	& WPE+BFIt & $\textbf{67.2}$ ($\textbf{58.7}$) \\
	\hline
	CNN-TDNNF	  									& None / $316$ 	&  BFIt & $68.7$ ($61.3$)  \\
	\hline
\end{tabular}
\end{center}
\label{tab:WERs_AM}
\end{table}

ASR results are given in Table~\ref{tab:WERs_AM}.
The first two rows show that  replacing the \gls{TDNN-F} with the CNN-TDNNF AM yielded more than \SI{2}{\%} absolute \gls{WER} reduction.
We also trained another CNN-TDNNF model using only a small subset (worn + 100k utterances from arrays) of training data (about \SI{316}{hrs} in total) which has produced slightly better \gls{WER}s compared with the baseline \gls{TDNN-F} trained on a much larger dataset (roughly \SI{1416}{hrs} in total). For consistency, 2-stage decoding was used for all results in Table~\ref{tab:WERs_AM}.
We conclude that the CNN-TDNNF model outperforms the TDNNF model for the CHiME-5 scenario and, therefore, for the remainder of the paper we only report results using the CNN-TDNNF AM.

\subsection{Enhancement effectiveness for ASR training and test}

An extensive set of experiments was performed to measure the WER impact of enhancement on the CHiME-5 training and test data. 
We test enhancement methods of varying strengths, as described in Section~\ref{ssec:setup}, and the results are depicted in Table~\ref{tab:WERs_enh}. 
In all cases, the (unprocessed) worn dataset was also included for AM training since it was found to improve performance (supporting therefore the argument that data variability helps ASR robustness).

\begin{table}[t]
\footnotesize
\caption{\gls{WER} results on the DEV (EVAL) set and various combinations of speech enhancement for ASR training and test (CNN-TDNNF AM). Amount of training data (\si{hrs}) is also specified.}
\begin{center}
	\setlength{\tabcolsep}{5pt}  
\begin{tabular}{lcccc}
	\hline
	 \multirow{2}{*}{\shortstack{Enh. in trng \\{ (\si{hrs})}}} 	& 	\multicolumn{4}{c}{Enhancement in test} \\ \cline{2-5}
																			& None & BFIt & GSS1 & GSS6 \\
	\hline
	None ($2046$)				& $69.3$ ($59.9$) & $69.1$ ($59.7$) & $62.2$ ($58.2$) & $51.8$ ($51.6$) \\
	BFIt ($680$)				& $68.9$ ($59.1$) & $68.5$ ($58.5$) & $59.9$ ($57.3$) & $48.8$ ($49.9$) \\
	GSS1 ($791$)				& $74.3$ ($67.5$) & $73.7$ ($66.4$) & $53.0$ ($49.6$) & $48.0$ ($47.5$) \\
	GSS6 ($308$)				& $78.5$ ($73.1$) & $76.9$ ($69.2$) & $58.0$ ($56.1$) & $\textbf{45.4}$ ($\textbf{45.7}$) \\

	\hline
\end{tabular}
\end{center}
\label{tab:WERs_enh}
\vspace{-10pt}
\end{table}

In Table~\ref{tab:WERs_enh}, in each row the recognition accuracy improves monotonically from left to right, i.e., as the enhancement strategy on the test data becomes stronger. Reading the table in each column from top to bottom, one observes that accuracy improves with increasing power of the enhancement on the training data, however, only as long as the enhancement on the training data is not stronger than on the test data.
Compared with unprocessed training and test data (None-None), GSS6-GSS6 yields roughly \SI{35}{\%} (\SI{24}{\%}) relative WER reduction on the DEV (EVAL) set, and \SI{12}{\%} (\SI{11}{\%}) relative WER reduction when compared with the None-GSS6 scenario. Comparing the amount of training data used to train the acoustic models, we observe that it decreases drastically from no enhancement to the GSS6 enhancement.

\subsection{State-of-the-art single-system for CHiME-5}

To facilitate comparison with the recently published top-line in~\cite{Kanda2019} (H/UPB), we have conducted a more focused set of experiments whose results are depicted in Table~\ref{tab:WERs_best}.
As explained in Section~\ref{sec:tempCtxtConfig}, we opted for~\cite{Kanda2019} instead of~\cite{Boeddecker2018chime5} as baseline because the former system is stronger.
The experiments include refining the GSS enhancement using time annotations from ASR output (GSS w/ ASR), performing discriminative training on top of the AMs trained with LF-MMI and performing RNN LM rescoring.
All the above helped further improve ASR performance. We report performance of our system on both single and multiple array tracks. To have a fair comparison, the results are compared with the single-system performance
reported in\cite{Kanda2019}.

For the \emph{single array track}, the proposed system without RNN LM rescoring achieves \SI{16}{\%} (\SI{11}{\%}) relative WER reduction on the DEV (EVAL) set when compared with System8 in~\cite{Kanda2019} (row one in Table~\ref{tab:WERs_best}). 
RNN LM rescoring further helps improve the proposed system performance. 

For the \emph{multi array track}, the proposed system without RNN LM rescoring achieved \SI{6}{\%} (\SI{7}{\%}) relative WER reduction on the DEV (EVAL) set when compared with System16 in~\cite{Kanda2019} (row six in Table~\ref{tab:WERs_best}). 

\begin{table}[t]
\footnotesize
\caption{Comparison of reference~\cite{Kanda2019} and proposed (single) systems in terms of \gls{WER} for the DEV (EVAL) set. Test data enhancement was refined using ASR alignments or oracle alignments.}
\begin{center}
	\setlength{\tabcolsep}{1.2pt}  
\begin{tabular}{llC{1.0cm}C{1.8cm}ccc}
	\hline
	
	Track & System & Enh. in trng & Enh. in \newline test & DT & RNN-LM & WER (\si{\%}) \\
	\hline
	
	\multirow{5}{*}{Single} & H/UPB~\cite{Kanda2019} & None & GSS1 w/ ASR & $\checkmark$ & & $58.3$ ($53.1$) \\
	
													& Proposed 	& GSS1 & GSS1 w/ ASR &  & & $50.2$ ($48.4$) \\
													
													& Proposed 	& GSS1 & GSS1 w/ ASR &  $\checkmark$ & & $49.1$ ($47.3$) \\
													
													& Proposed 	& GSS1 & GSS1 w/ ASR & $\checkmark$ & $\checkmark$ & $\bf{48.6}$ ($\bf{46.7}$) \\ \cline{2-7}
													
													& \textcolor{gray}{Proposed} & \textcolor{gray}{GSS1} & \textcolor{gray}{GSS1 w/ oracle} & \textcolor{gray}{$\checkmark$}  & \textcolor{gray}{$\checkmark$} & \textcolor{gray}{$47.3$ ($46.1$)} \\
													\hline
	
	\multirow{5}{*}{Multiple} & H/UPB~\cite{Kanda2019} & None & GSS6 w/ ASR & $\checkmark$ & & $45.1$ ($47.3$) \\
	
													& Proposed 	& GSS6 & GSS6 w/ ASR &  & & $43.2$ ($44.2$) \\
													
													& Proposed 	& GSS6 & GSS6 w/ ASR & $\checkmark$ & & $42.3$ ($43.9$) \\
													
													& Proposed 	& GSS6 & GSS6 w/ ASR & $\checkmark$ & $\checkmark$ & $\bf{41.6}$ ($\bf{43.2}$) \\ \cline{2-7}
													
													& \textcolor{gray}{Proposed} 	& \textcolor{gray}{GSS6} & \textcolor{gray}{GSS6 w/ oracle} & \textcolor{gray}{$\checkmark$} & \textcolor{gray}{$\checkmark$} & \textcolor{gray}{$39.9$ ($42.0$)} \\
													\hline
\end{tabular}
\end{center}
\label{tab:WERs_best}
\end{table}

We also performed a test using GSS with the oracle alignments (GSS w/ oracle) to assess the potential of time annotation refinement  (gray shade lines in Table~\ref{tab:WERs_best}). It can be seen that there is some, however not much room for improvement.

Finally, cleaning up the training set not only boosted the recognition performance, 
but managed to do so using a fraction of the training data in~\cite{Kanda2019}, as shown in Table~\ref{tab:WERs_hrs}.
This translates to significantly faster and cheaper training of acoustic models, which is a major advantage in practice.

\begin{table}[t]
\footnotesize
\caption{Comparison of the reference~\cite{Kanda2019} and proposed systems in terms of amount of training data.}

\begin{center}
	\setlength{\tabcolsep}{3pt}  
\begin{tabular}{llcc}
	\hline
	
	Track & System & Amount trng data (\si{hrs}) & WER (\si{\%}) \\
	\hline
	
	\multirow{2}{*}{Single} & H/UPB~\cite{Kanda2019} & $4500$ & $58.3$ ($53.1$) \\
													
													& Proposed 	& $791$ & $\bf{48.6}$ ($\bf{46.7}$) \\
													
													\hline
	
	\multirow{2}{*}{Multiple} & H/UPB~\cite{Kanda2019} & $4500$ & $45.1$ ($47.3$) \\
													
													& Proposed 	& $308$ & $\bf{41.6}$ ($\bf{43.2}$) \\
													
													\hline
\end{tabular}
\end{center}
\label{tab:WERs_hrs}
\vspace{-10pt}
\end{table}

\section{Discussion}
\label{sec:disc}

\subsection{Temporal context configuration for GSS}
\label{sec:tempCtxtConfig}

Our experiments have shown that the temporal context of some GSS components has a significant effect on the WER. Two cases are investigated: (i) partially dropping the temporal context for the EM stage, and (ii) dropping the temporal context for beamforming.
The evaluation was conducted with an acoustic model trained on unprocessed speech and the enhancement was applied during test only.
Results are depicted in Table~\ref{tab:WERs_context}.

The first row corresponds to the GSS configuration in~\cite{Boeddecker2018chime5} while the second one corresponds to the GSS configuration in~\cite{Kanda2019}. First two rows show
that dropping the temporal context for estimating statistics for beamforming improves ASR accuracy.
For the last row, the EM algorithm was run 20 iterations with temporal context, followed by another 10 without context. Since the performance decreased, we concluded that the best configuration for the \gls{GSS} enhancement in CHiME-5 scenario is using full temporal context for the EM stage and dropping it for the beamforming stage. Consequently, we have chosen system~\cite{Kanda2019} as baseline in this study since is using the stronger GSS configuration.

\begin{table}[t]
\footnotesize
\caption{WER results using CNN-TDNNF AM trained on unprocessed (None) when some \gls{GSS} enhancement (test) components ignore the temporal context.}

\begin{center}
\begin{tabular}{llc}
	\hline
	EM iterations & BF & WER (\si{\%})\\
	\hline
	20 w/ context~\cite{Boeddecker2018chime5} & w/ context & $54.7$ ($52.3$) \\
	20 w/ context~\cite{Kanda2019} 						& w/o context& $\bf{51.8}$ ($\bf{51.6}$) \\
	20 w/ + 10 w/o context 										& w/o context & $52.2$ ($52.5$) \\
	\hline
\end{tabular}
\end{center}
\label{tab:WERs_context}
\vspace{-10pt}
\end{table}

\subsection{Analysis of speaker overlap effect on WER accuracy}

\begin{figure}[b!]
	\begin{centering}
		\begin{tikzpicture}[font=\footnotesize]

\pgfplotsset{
	nodes near coords always on top/.style={
		scatter/position=absolute,
		positive value/.style={
			at={(axis cs:\pgfkeysvalueof{/data point/x},\pgfkeysvalueof{/data point/y})},
		},
		negative value/.style={
			at={(axis cs:\pgfkeysvalueof{/data point/x},0)},
		},
		every node near coord/.append style={
			check values/.code={%
				\begingroup
				\pgfkeys{/pgf/fpu}%
				\pgfmathparse{\pgfplotspointmeta<0}%
				\global\let\result=\pgfmathresult
				\endgroup
				%
				%
				\pgfmathfloatcreate{1}{1.0}{0}%
				\let\ONE=\pgfmathresult
				\ifx\result\ONE
				\pgfkeysalso{/pgfplots/negative value}%
				\else
				\pgfkeysalso{/pgfplots/positive value}%
				\fi
			},
			check values,
			anchor=west,
			rotate=90,
			fill=white,
			inner ysep=1,
			inner xsep=1,
			outer xsep=3,
			outer ysep=1,
		},
	},
}

\begin{axis}[
	width  = 0.97*\columnwidth,
	height = 5cm,
	major x tick style = transparent,
	ybar=5*\pgflinewidth,
	ymin=0, ymax=0.9,
	ytick={0.0,0.1,...,0.91},
	bar width=9pt,
	ymajorgrids = true,
	ylabel near ticks,
	axis line style={opacity=0},
	xlabel = {Amount of speaker overlap (\%)},
	ylabel = {Word frequency},
	ytick style={draw=none},
	symbolic x coords={$0-20$,$20-40$,$40-60$,$60-80$,$80-100$},
	xtick = data,
	scaled y ticks = false,
	enlarge x limits=0.13,
	legend cell align=left,
	legend style={
		at={(0.4,0.93)},
		anchor=center,
		legend columns=-1,
		/tikz/every even column/.append style={column sep=2ex},
		draw=none,
		fill=none,
	},
	nodes near coords=\rotatebox{0}{{\pgfmathprintnumber[fixed,precision=2]{\pgfplotspointmeta}}},
	nodes near coords always on top,
]

\pgfplotsset{
	legend image code/.code={
		\draw [#1] (0cm,-0.08cm) rectangle (0.2cm,0.12cm);
	},
}

\pgfplotstableread[row sep=\\,col sep=&]{
	overlap & dev & eval \\
	{$0-20$} & 0.117 & 0.343 \\  
	{$20-40$} & 0.162 & 0.164 \\
	{$40-60$} & 0.183 & 0.134 \\
	{$60-80$} & 0.150 & 0.105 \\
	{$80-100$} & 0.389 & 0.253 \\
}\mydata

\definecolor{bblue}{HTML}{4F81BD}
\definecolor{rred}{HTML}{C0504D}
\definecolor{ggreen}{HTML}{9BBB59}
\definecolor{ppurple}{HTML}{9F4C7C}

\addplot[style={bblue,fill=bblue,mark=none}] table[x=overlap,y=dev]{\mydata};
\addplot[style={rred,fill=rred,mark=none}] table[x=overlap,y=eval]{\mydata};

\legend{{DEV},{EVAL}}
\end{axis}
\end{tikzpicture}
		\caption{Word distribution of overlapped speech for the DEV and EVAL sets of CHiME-5.}
		\label{fig:data_distr}
	\end{centering}
\end{figure}
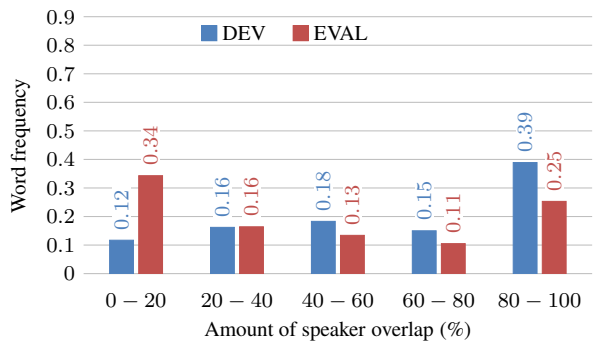

The results presented so far were overall accuracies on the test set of CHiME-5.
However, since speaker overlap is a major issue for these data, it is of interest to investigate the methods' performance as a function of the amount of overlapped speech. Employing the original CHiME-5 annotations, the word distribution of overlapped speech was computed for DEV and EVAL sets (silence portions were not filtered out).
The five-bin normalized histogram of the data is plotted in Fig.~\ref{fig:data_distr}.
Interestingly, the percentage of segments with low overlapped speech is significantly higher for the EVAL than for the DEV set, and, conversely, the number of words with high overlapped speech is considerably lower for the EVAL than for the DEV set.
This distribution may explain the difference in performance observed between the DEV and EVAL sets.

\begin{table}[t]
\footnotesize
\caption{Breakdown of absolute \gls{WER} results on the DEV (EVAL) set for the same training and test enhancement (matched case, CNN-TDNNF AM).}
\begin{center}
	\setlength{\tabcolsep}{1.5pt}  
\begin{tabular}{lccccc}
	\hline
	 \multirow{2}{*}{\shortstack{Enh. \\(trng+test)}} 	& 	\multicolumn{5}{c}{Amount of overlap (\si{\%})} \\ \cline{2-6}
																			& $0-20$ & $20-40$ & $40-60$ & $60-80$ & $80-100$ \\
	\hline
	None				& $48.3$ ($47.2$) & $49.0$ ($49.5$) & $56.9$ ($57.4$) & $64.5$ ($67.4$) & $89.5$ ($84.1$) \\
	BFIt				& $46.6$ ($45.6$) & $47.8$ ($48.4$) & $54.9$ ($55.0$) & $63.6$ ($66.7$) & $89.5$ ($84.2$) \\
	GSS1				& $42.2$ ($43.3$) & $41.6$ ($43.4$) & $44.8$ ($47.8$) & $50.6$ ($55.3$) & $69.0$ ($67.6$) \\
	GSS6				& $36.5$ ($40.1$) & $36.4$ ($40.8$) & $41.0$ ($44.6$) & $43.8$ ($49.9$) & $58.8$ ($62.0$) \\

	\hline
\end{tabular}
\end{center}
\label{tab:WER_overlap_matched}
\end{table}

\begin{table}[t]
\footnotesize
\caption{Breakdown of absolute \gls{WER} results on the DEV (EVAL) set for unprocessed training data and various test enhancements (mismatched case, CNN-TDNNF AM).}
\begin{center}
	\setlength{\tabcolsep}{1pt}  
\begin{tabular}{L{1.5cm}ccccc}
	\hline
	 \multirow{2}{*}{\shortstack{Enh. (test)}} 	& 	\multicolumn{5}{c}{Amount of overlap (\si{\%})} \\ \cline{2-6}
																			& $0-20$ & $20-40$ & $40-60$ & $60-80$ & $80-100$ \\
	\hline
	None				& $48.3$ ($47.2$) & $49.0$ ($49.5$) & $56.9$ ($57.4$) & $64.5$ ($67.4$) & $89.5$ ($84.1$) \\
	BFIt				& $47.5$ ($47.0$) & $48.4$ ($49.7$) & $56.5$ ($56.6$) & $64.3$ ($66.9$) & $89.3$ ($83.7$) \\
	GSS1				& $48.8$ ($51.1$) & $49.2$ ($51.4$) & $53.4$ ($55.3$) & $58.5$ ($63.5$) & $78.3$ ($76.2$) \\
	GSS1 \newline ~~ w/o Mask& $44.0$ ($44.9$) & $45.8$ ($46.8$) & $51.5$ ($52.9$) & $57.7$ ($62.4$) & $82.4$ ($78.2$) \\
	GSS6				& $40.3$ ($45.5$) & $41.2$ ($45.1$) & $45.1$ ($50.0$) & $48.2$ ($54.9$) & $66.7$ ($68.9$) \\
	GSS6 \newline ~~ w/ ASR			& $38.8$ ($44.5$) & $39.8$ ($43.8$) & $43.3$ ($49.2$) & $46.4$ ($53.4$) & $63.5$ ($67.1$) \\

	\hline
\end{tabular}
\end{center}
\label{tab:WER_overlap_mismatched}
\vspace{-15pt}
\end{table}

Based on the distributions in Fig.~\ref{fig:data_distr}, the test data was split. Two cases were considered: (a) same enhancement for training and test data (matched case, Table~\ref{tab:WER_overlap_matched}), and (b) unprocessed training data and enhanced test data (mismatched case, Table~\ref{tab:WER_overlap_mismatched}). As expected, the WER increases monotonically as the amount of overlap increases in both scenarios, and the recognition accuracy improves as the enhancement method becomes stronger.

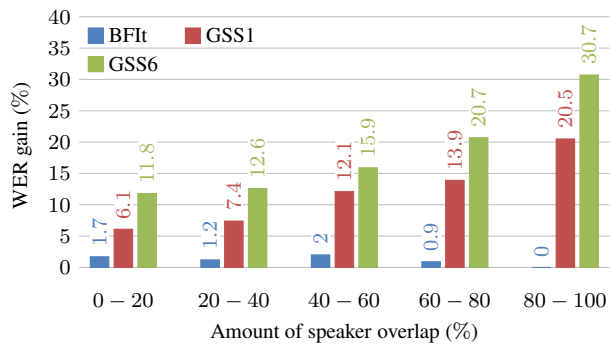
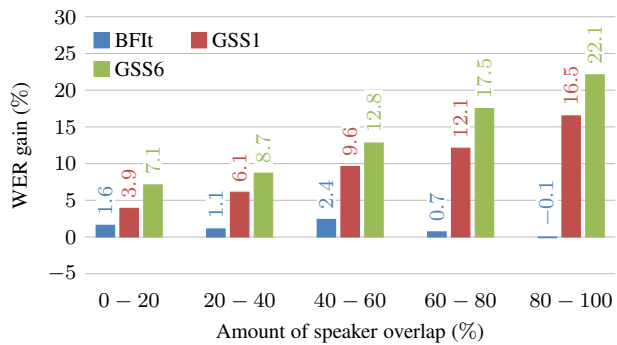
\begin{figure*}[t]
	\begin{subfigure}{0.5\textwidth}\centering
		\begin{tikzpicture}[font=\footnotesize]

\pgfplotsset{
	nodes near coords always on top/.style={
		scatter/position=absolute,
		positive value/.style={
			at={(axis cs:\pgfkeysvalueof{/data point/x},\pgfkeysvalueof{/data point/y})},
		},
		negative value/.style={
			at={(axis cs:\pgfkeysvalueof{/data point/x},0)},
		},
		every node near coord/.append style={
			check values/.code={%
				\begingroup
				\pgfkeys{/pgf/fpu}%
				\pgfmathparse{\pgfplotspointmeta<0}%
				\global\let\result=\pgfmathresult
				\endgroup
				%
				%
				\pgfmathfloatcreate{1}{1.0}{0}%
				\let\ONE=\pgfmathresult
				\ifx\result\ONE
				\pgfkeysalso{/pgfplots/negative value}%
				\else
				\pgfkeysalso{/pgfplots/positive value}%
				\fi
			},
			check values,
			anchor=west,
			rotate=90,
			fill=white,
			inner ysep=1,
			inner xsep=1,
			outer xsep=3,
			outer ysep=1,
		},
	},
}

\begin{axis}[
	width  = 0.97*\textwidth,
	height = 5cm,
	major x tick style = transparent,
	ybar=5*\pgflinewidth,
	ymin=-1, ymax=40,
	ytick={0,5,...,40}, 
	bar width=7pt,
	ymajorgrids = true,
	ylabel near ticks,
	axis line style={opacity=0},
	xlabel = {Amount of speaker overlap (\si{\%})},
	ylabel = {WER gain (\si{\%})},
	ytick style={draw=none},
	symbolic x coords={$0-20$,$20-40$,$40-60$,$60-80$,$80-100$},
	xtick = data,
	scaled y ticks = false,
	legend cell align=left,
	legend style={
		at={(0.00,0.999)},
		anchor=north west,
		legend columns=2,
		/tikz/every even column/.append style={column sep=2ex},
		draw=none,
		fill=none,
	},
	nodes near coords=\rotatebox{0}{{\pgfmathprintnumber[fixed,precision=1]{\pgfplotspointmeta}}},
	nodes near coords always on top,
]

\pgfplotsset{
	legend image code/.code={
		\draw [#1] (0cm,-0.08cm) rectangle (0.2cm,0.12cm);
	},
}

\pgfplotstableread[row sep=\\,col sep=&]{
	overlap & beamformIt & gss1 & gss6 \\
	{$0-20$}    & 1.7 & 6.1 & 11.8 \\
	{$20-40$}    & 1.2 & 7.4 & 12.6 \\
	{$40-60$}    & 2 & 12.1 & 15.9 \\
	{$60-80$}    & 0.9 & 13.9 & 20.7 \\
	{$80-100$}    & 0 & 20.5 & 30.7 \\
}\mydata

\definecolor{bblue}{HTML}{4F81BD}
\definecolor{rred}{HTML}{C0504D}
\definecolor{ggreen}{HTML}{9BBB59}
\definecolor{ppurple}{HTML}{9F4C7C}

\addplot[style={bblue,fill=bblue,mark=none}] table[x=overlap,y=beamformIt]{\mydata};
\addplot[style={rred,fill=rred,mark=none}] table[x=overlap,y=gss1]{\mydata};
\addplot[style={ggreen,fill=ggreen,mark=none}] table[x=overlap,y=gss6]{\mydata};

\legend{{BFIt},{GSS1},{GSS6}}
\end{axis}
\end{tikzpicture}
		\caption{DEV}
		\label{fig:matched_dev}
	\end{subfigure}
	\begin{subfigure}{0.5\textwidth}\centering
		\begin{tikzpicture}[font=\footnotesize]

\pgfplotsset{
	nodes near coords always on top/.style={
		scatter/position=absolute,
		positive value/.style={
			at={(axis cs:\pgfkeysvalueof{/data point/x},\pgfkeysvalueof{/data point/y})},
		},
		negative value/.style={
			at={(axis cs:\pgfkeysvalueof{/data point/x},0)},
		},
		every node near coord/.append style={
			check values/.code={%
				\begingroup
				\pgfkeys{/pgf/fpu}%
				\pgfmathparse{\pgfplotspointmeta<0}%
				\global\let\result=\pgfmathresult
				\endgroup
				%
				%
				\pgfmathfloatcreate{1}{1.0}{0}%
				\let\ONE=\pgfmathresult
				\ifx\result\ONE
				\pgfkeysalso{/pgfplots/negative value}%
				\else
				\pgfkeysalso{/pgfplots/positive value}%
				\fi
			},
			check values,
			anchor=west,
			rotate=90,
			fill=white,
			inner ysep=1,
			inner xsep=1,
			outer xsep=3,
			outer ysep=1,
		},
	},
}

\begin{axis}[
	width  = 0.97*\textwidth,
	height = 5cm,
	major x tick style = transparent,
	ybar=5*\pgflinewidth,
	ymin=-5, ymax=30,
	ytick={-5,0,...,40}, 
	bar width=7pt,
	ymajorgrids = true,
	ylabel near ticks,
	axis line style={opacity=0},
	xlabel = {Amount of speaker overlap (\si{\%})},
	ylabel = {WER gain (\si{\%})},
	ytick style={draw=none},
	symbolic x coords={$0-20$,$20-40$,$40-60$,$60-80$,$80-100$},
	xtick = data,
	scaled y ticks = false,
	legend cell align=left,
	legend style={
		at={(0.00,0.98)},
		anchor=north west,
		legend columns=2,
		/tikz/every even column/.append style={column sep=2ex},
		draw=none,
		fill=none,
	},
	nodes near coords=\rotatebox{0}{{\pgfmathprintnumber[fixed,precision=1]{\pgfplotspointmeta}}},
	nodes near coords always on top,
]

\pgfplotsset{
	legend image code/.code={
		\draw [#1] (0cm,-0.08cm) rectangle (0.2cm,0.12cm);
	},
}

\pgfplotstableread[row sep=\\,col sep=&]{
	overlap & beamformIt & gss1 & gss6 \\
	{$0-20$}    & 1.6 & 3.9  & 7.1  \\    
	{$20-40$}   & 1.1 & 6.1  & 8.7  \\    
	{$40-60$}   & 2.4 & 9.6  & 12.8 \\    
	{$60-80$}   & 0.7 & 12.1   & 17.5 \\    
	{$80-100$}  & -0.1 & 16.5 & 22.1 \\    
}\mydata

\definecolor{bblue}{HTML}{4F81BD}
\definecolor{rred}{HTML}{C0504D}
\definecolor{ggreen}{HTML}{9BBB59}
\definecolor{ppurple}{HTML}{9F4C7C}

\addplot[style={bblue,fill=bblue,mark=none}] table[x=overlap,y=beamformIt]{\mydata};
\addplot[style={rred,fill=rred,mark=none}] table[x=overlap,y=gss1]{\mydata};
\addplot[style={ggreen,fill=ggreen,mark=none}] table[x=overlap,y=gss6]{\mydata};

\legend{{BFIt},{GSS1},{GSS6}}
\end{axis}
\end{tikzpicture}
		\caption{EVAL}
		\label{fig:matched_eval}
	\end{subfigure}
\caption{Relative WER gain for the matched case vs unprocessed, Table~\ref{tab:WER_overlap_matched} row one (CNN-TDNNF AM).}
\label{fig:matched}
\end{figure*}

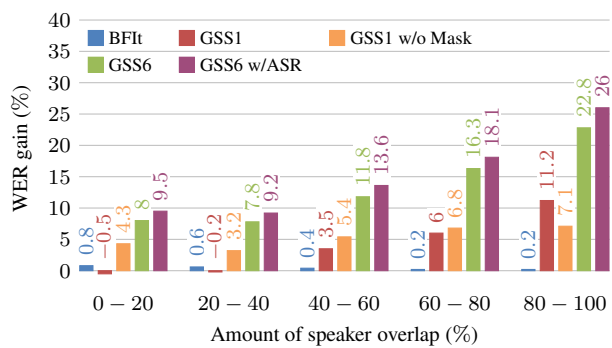
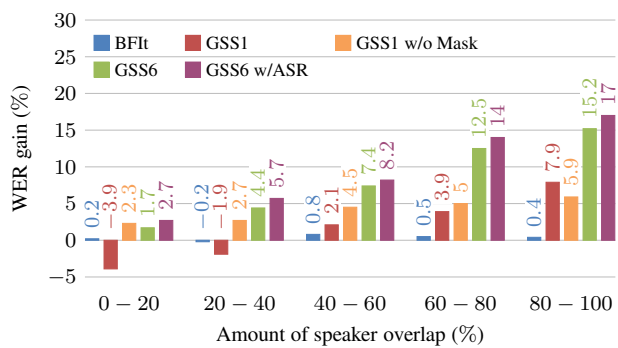
\begin{figure*}[t]
	\begin{subfigure}{0.5\textwidth}\centering
		\begin{tikzpicture}[font=\footnotesize]

\pgfplotsset{
	nodes near coords always on top/.style={
		scatter/position=absolute,
		positive value/.style={
			at={(axis cs:\pgfkeysvalueof{/data point/x},\pgfkeysvalueof{/data point/y})},
		},
		negative value/.style={
			at={(axis cs:\pgfkeysvalueof{/data point/x},0)},
		},
		every node near coord/.append style={
			check values/.code={%
				\begingroup
				\pgfkeys{/pgf/fpu}%
				\pgfmathparse{\pgfplotspointmeta<0}%
				\global\let\result=\pgfmathresult
				\endgroup
				%
				%
				\pgfmathfloatcreate{1}{1.0}{0}%
				\let\ONE=\pgfmathresult
				\ifx\result\ONE
				\pgfkeysalso{/pgfplots/negative value}%
				\else
				\pgfkeysalso{/pgfplots/positive value}%
				\fi
			},
			check values,
			anchor=west,
			rotate=90,
			fill=white,
			inner ysep=1,
			inner xsep=1,
			outer xsep=3,
			outer ysep=1,
		},
	},
}

\begin{axis}[
	width  = 0.97*\textwidth,
	height = 5cm,
	major x tick style = transparent,
	ybar=5*\pgflinewidth,
	ymin=-1, ymax=40,
	ytick={-5,0,...,40},
	bar width=5pt,
	ymajorgrids = true,
	ylabel near ticks,
	axis line style={opacity=0},
	xlabel = {Amount of speaker overlap (\si{\%})},
	ylabel = {WER gain (\si{\%})},
	ytick style={draw=none},
	symbolic x coords={$0-20$,$20-40$,$40-60$,$60-80$,$80-100$},
	xtick = data,
	scaled y ticks = false,
	legend cell align=left,
	legend style={
		at={(0.00,0.999)},
		anchor=north west,
		legend columns=3,
		/tikz/every even column/.append style={column sep=2ex},
		draw=none,
		fill=none,
	},
	nodes near coords=\rotatebox{0}{{\pgfmathprintnumber[fixed,precision=1]{\pgfplotspointmeta}}},
	nodes near coords always on top,
]

\pgfplotsset{
	legend image code/.code={
		\draw [#1] (0cm,-0.08cm) rectangle (0.2cm,0.12cm);
	},
}

\pgfplotstableread[row sep=\\,col sep=&]{
	overlap & beamformIt & gss1 & gsswoMask1 & gss6 & gssASR6 \\
	{$0-20$}    & 0.8 & -0.5 	& 4.3 & 8    & 9.5  \\  
	{$20-40$}   & 0.6 & -0.2 	& 3.2 & 7.8  & 9.2  \\  
	{$40-60$}   & 0.4 & 3.5	& 5.4 & 11.8 & 13.6 \\  
	{$60-80$}   & 0.2 & 6   	& 6.8 & 16.3 & 18.1 \\  
	{$80-100$}  & 0.2 & 11.2 	& 7.1 & 22.8 & 26   \\  
}\mydata

\definecolor{bblue}{HTML}{4F81BD}
\definecolor{rred}{HTML}{C0504D}
\definecolor{ggreen}{HTML}{9BBB59}
\definecolor{ppurple}{HTML}{9F4C7C} 
\definecolor{oorange}{HTML}{f7a057}

\addplot[style={bblue,fill=bblue,mark=none}] table[x=overlap,y=beamformIt]{\mydata};
\addplot[style={rred,fill=rred,mark=none}] table[x=overlap,y=gss1]{\mydata};
\addplot[style={oorange,fill=oorange,mark=none}] table[x=overlap,y=gsswoMask1]{\mydata};
\addplot[style={ggreen,fill=ggreen,mark=none}] table[x=overlap,y=gss6]{\mydata};
\addplot[style={ppurple,fill=ppurple,mark=none}] table[x=overlap,y=gssASR6]{\mydata};

\legend{\scriptsize{BFIt},\scriptsize{GSS1},\scriptsize{GSS1 w/o Mask},\scriptsize{GSS6},\scriptsize{GSS6 w/ASR}}
\end{axis}
\end{tikzpicture}
		\caption{DEV}
		\label{fig:mmatchedA_dev}
	\end{subfigure}
	\begin{subfigure}{0.5\textwidth}\centering
		\begin{tikzpicture}[font=\footnotesize]

\pgfplotsset{
	nodes near coords always on top/.style={
		scatter/position=absolute,
		positive value/.style={
			at={(axis cs:\pgfkeysvalueof{/data point/x},\pgfkeysvalueof{/data point/y})},
		},
		negative value/.style={
			at={(axis cs:\pgfkeysvalueof{/data point/x},0)},
		},
		every node near coord/.append style={
			check values/.code={%
				\begingroup
				\pgfkeys{/pgf/fpu}%
				\pgfmathparse{\pgfplotspointmeta<0}%
				\global\let\result=\pgfmathresult
				\endgroup
				%
				%
				\pgfmathfloatcreate{1}{1.0}{0}%
				\let\ONE=\pgfmathresult
				\ifx\result\ONE
				\pgfkeysalso{/pgfplots/negative value}%
				\else
				\pgfkeysalso{/pgfplots/positive value}%
				\fi
			},
			check values,
			anchor=west,
			rotate=90,
			fill=white,
			inner ysep=1,
			inner xsep=1,
			outer xsep=3,
			outer ysep=1,
		},
	},
}

\begin{axis}[
	width  = 0.97*\textwidth,
	height = 5cm,
	major x tick style = transparent,
	ybar=5*\pgflinewidth,
	ymin=-5, ymax=30,
	ytick={-5,0,...,40}, 
	bar width=5pt,
	ymajorgrids = true,
	ylabel near ticks,
	axis line style={opacity=0},
	xlabel = {Amount of speaker overlap (\si{\%})},
	ylabel = {WER gain (\si{\%})},
	ytick style={draw=none},
	symbolic x coords={$0-20$,$20-40$,$40-60$,$60-80$,$80-100$},
	xtick = data,
	scaled y ticks = false,
	legend cell align=left,
	legend style={
		at={(0.00,0.98)},
		anchor=north west,
		legend columns=3,
		/tikz/every even column/.append style={column sep=2ex},
		draw=none,
		fill=none,
	},
	nodes near coords=\rotatebox{0}{{\pgfmathprintnumber[fixed,precision=1]{\pgfplotspointmeta}}},
	nodes near coords always on top,
]

\pgfplotsset{
	legend image code/.code={
		\draw [#1] (0cm,-0.08cm) rectangle (0.2cm,0.12cm);
	},
}

\pgfplotstableread[row sep=\\,col sep=&]{
	overlap & beamformIt & gss1 & gsswoMask1 & gss6 & gssASR6 \\
	{$0-20$}   & 0.2  & -3.9 & 2.3 & 1.7  & 2.7  \\ 
	{$20-40$}  & -0.2 & -1.9 & 2.7  & 4.4  & 5.7  \\ 
	{$40-60$}  & 0.8  & 2.1  & 4.5  & 7.4  & 8.2  \\ 
	{$60-80$}  & 0.5  & 3.9 & 5.0  & 12.5 & 14.0 \\ 
	{$80-100$} & 0.4  & 7.9 & 5.9  & 15.2 & 17.0 \\ 
}\mydata

\definecolor{bblue}{HTML}{4F81BD}
\definecolor{rred}{HTML}{C0504D}
\definecolor{ggreen}{HTML}{9BBB59}
\definecolor{ppurple}{HTML}{9F4C7C}
\definecolor{oorange}{HTML}{f7a057}

\addplot[style={bblue,fill=bblue,mark=none}] table[x=overlap,y=beamformIt]{\mydata};
\addplot[style={rred,fill=rred,mark=none}] table[x=overlap,y=gss1]{\mydata};
\addplot[style={oorange,fill=oorange,mark=none}] table[x=overlap,y=gsswoMask1]{\mydata};
\addplot[style={ggreen,fill=ggreen,mark=none}] table[x=overlap,y=gss6]{\mydata};
\addplot[style={ppurple,fill=ppurple,mark=none}] table[x=overlap,y=gssASR6]{\mydata};

\legend{\scriptsize{BFIt},\scriptsize{GSS1},\scriptsize{GSS1 w/o Mask},\scriptsize{GSS6},\scriptsize{GSS6 w/ASR}}
\end{axis}
\end{tikzpicture}
		\caption{EVAL}
		\label{fig:mmatchedA_eval}
	\end{subfigure}
\caption{Relative WER gain for the mismatched case vs unprocessed, Table~\ref{tab:WER_overlap_mismatched} row one (CNN-TDNNF AM trained on unprocessed).}
\label{fig:mmmatchedA}
\end{figure*}

Graphical representations of WER gains (relative to the unprocessed case) in Tables~\ref{tab:WER_overlap_matched} and \ref{tab:WER_overlap_mismatched} are given in Figs.~\ref{fig:matched} and \ref{fig:mmmatchedA}.
The plots show that as the amount of speaker overlap increases, the accuracy gain (relative to the unprocessed case) of the weaker signal enhancement (BFIt) drops. 
This is an expected result since BFIt is not a source separation algorithm.
Conversely, as the amount of speaker overlap increases, the accuracy gain (relative to None) of the stronger GSS enhancement improves quite significantly.
A rather small decrease in accuracy is observed in the mismatched case (Fig.~\ref{fig:mmmatchedA}) for GSS1 in the lower overlap regions.
As already mentioned in Section~\ref{sec:gss}, this is due to the masking stage. It has previously been observed that using masking for speech enhancement without a cross talker decreases ASR recognition performance. 
We have also included in Fig.~\ref{fig:mmmatchedA} the GSS1 version without masking (GSS w/o Mask), which indeed yields significant accuracy gains on segments with little overlap. 
However, since the overall accuracy of GSS1 with masking is higher than the overall gain of GSS1 without masking, GSS w/o mask was not included in the previous experiments.

\section{Conclusions}
\label{sec:concl}

In this paper we performed an extensive experimental evaluation on the acoustically very challenging CHiME-5 dinner party data showing that:
(i) cleaning up training data can lead to substantial word error rate reduction, and (ii) enhancement in training is advisable as long as enhancement in test is at least as strong as in training.
This approach stands in contrast and delivers larger accuracy gains at a fraction of training data than the common data simulation strategy found in the literature.
Using a CNN-TDNNF acoustic model topology along with GSS enhancement refined with time annotations from ASR, discriminative training and RNN LM rescoring, we achieved a new single-system state-of-the-art result on CHiME-5, which is \SI{41.6}{\%} (\SI{43.2}{\%}) on the development (evaluation) set, which is a \SI{8}{\%} relative improvement of the word error rate over a comparable system reported so far.

\section{Acknowledgments}

Parts of computational resources required in this study were provided by the Paderborn Center for Parallel Computing.

\newpage

\balance
\bibliographystyle{myIEEEbib}
\bibliography{IEEEabrv,refs}

\end{document}